\documentclass[10pt,twocolumn,letterpaper]{article}
\usepackage{cvpr}

\usepackage{algorithm}
\usepackage{algorithmic}
\usepackage{array}
\usepackage{tabularx}
\newcolumntype{Y}{>{\centering\arraybackslash}X}

\definecolor{cvprblue}{rgb}{0.21,0.49,0.74}
\usepackage[breaklinks,colorlinks,allcolors=cvprblue]{hyperref}
\def\confName{CVPR}
\def\confYear{2027}
\usepackage{silence}
\usepackage{microtype}
\AtBeginDocument{\lefthyphenmin=2\righthyphenmin=3}

\newcommand{\guiqdisplayspacing}{%
  \abovedisplayskip=3pt plus 1pt minus 1pt
  \belowdisplayskip=3pt plus 1pt minus 1pt
  \abovedisplayshortskip=0pt plus 1pt
  \belowdisplayshortskip=3pt plus 1pt minus 1pt
  \jot=2pt
}
\AtBeginEnvironment{equation}{\guiqdisplayspacing}
\AtBeginEnvironment{align}{\guiqdisplayspacing}
\AtBeginEnvironment{gather}{\guiqdisplayspacing}
\allowdisplaybreaks[1]

\renewcommand{\paragraph}[1]{\par\addvspace{.5em plus 1pt minus .5pt}\noindent\textbf{#1}\enspace\ignorespaces}

\title{GUI-Q: Task-Sensitive Granularity and Relation-Preserving Quantization \\ for GUI Agents}

\author{
    Tianqi Li\textsuperscript{1,*},
    Yang Wang\textsuperscript{2,*},
    Xin He\textsuperscript{3},
    Le Zhang\textsuperscript{4},
    Yun Liu\textsuperscript{1,5,6,\textdagger}\\
    {\small \textsuperscript{1}VCIP, College of Computer Science, Nankai University}\\
    {\small \textsuperscript{2}College of Mathematics and Physics, Beijing University of Chemical Technology}\\
    {\small \textsuperscript{3}School of Computer Science and Engineering, Tianjin University of Technology}\\
    {\small \textsuperscript{4}College of Computer Science and Software Engineering, Hohai University}\\
    {\small \textsuperscript{5}Academy for Advanced Interdisciplinary Studies, Nankai University}\\
    {\small \textsuperscript{6}Nankai International Advanced Research Institute, Shenzhen Futian}\\
    {\small \textsuperscript{*}Equal contribution. \textsuperscript{\textdagger}Corresponding author.}
}

\AtEndDocument{\typeout{GUIQ-FIG1-PAGE=\getpagerefnumber{fig:intro}}}

\begin{document}
\maketitle

\begin{figure*}[!t]
\centering
\includegraphics[width=\textwidth]{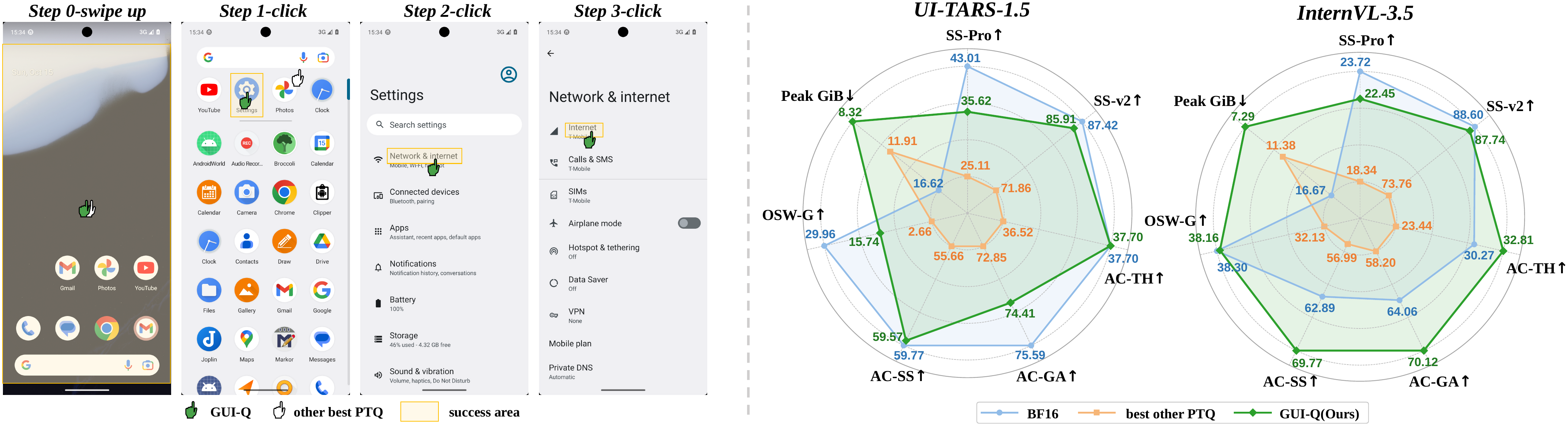}
\caption{Task and cross-model comparison. Left: four-step trajectory comparisons. Right: BF16, the strongest PTQ baselines, and GUI-Q across six accuracy metrics and peak memory.}
\label{fig:intro}
\end{figure*}

\begin{abstract}

\looseness=-1\noindent

Graphical user interface (GUI) agents enable automated interactions with mobile, web, and desktop interfaces. However, their large memory footprint and repeated model calls hinder deployment on resource-constrained devices, motivating post-training quantization (PTQ) without full retraining. Quantization can degrade GUI response accuracy and distort the instruction--vision affinities and local spatial relations needed for GUI target selection. Small coordinate errors can cause missed clicks, while early action errors can alter subsequent interface states and disrupt task execution. Local reconstruction error alone does not directly assess whether these responses and relations are preserved. We propose \textbf{GUI-Q}, a PTQ framework that targets response and relation preservation with 4-bit weights and 16-bit activations (W4A16). \textbf{Task-Sensitive Granularity Allocation (TSGA)} allocates finer groups under a scale-byte budget by combining response sensitivity with the measured reduction in activation-weighted distortion. \textbf{Relation-Preserving Symmetry Selection (RPSS)} then selects symmetric or asymmetric grids to preserve instruction--vision affinities and local spatial relations in GUIs. Across three GUI-capable models, GUI-Q improves mean grounding TargetHit by 8.04--12.55 points over the strongest evaluated 4-bit configurations, while reducing static peak graphics processing unit (GPU) memory by 49.9--56.2\% relative to bfloat16 (BF16). Fixed-output replay with packed 4-bit execution provides 1.7--2.8$\times$ model-call speedups. The code will be released.

\end{abstract}

\section{Introduction}

GUI agents translate instructions and screenshots into actions on mobile, web, and desktop interfaces. Beyond understanding images and language \cite{internvl35,qvlm,vlmptq}, they must localize targets precisely and specify action arguments correctly. Small coordinate errors can cause missed clicks, and early mistakes can alter the interface state in later steps.

Recent GUI agents \cite{uitars,osatlas,agentcpmgui} rely on large vision-language models and repeated model calls, creating substantial memory and inference costs. For 7–8B models, FP32 weights require approximately 26--30 GiB of memory. Even in bfloat16 (BF16), peak GPU memory allocation reaches 15.75--16.67 GiB in our static evaluation (Sec.~\ref{sec:efficiency}). These resource demands complicate local deployment on smartphones and other edge devices with limited memory and compute resources. Post-training quantization (PTQ) offers a practical route to reduce weight storage while reusing pretrained models without full retraining. We focus on four-bit weight quantization (W4) to reduce model cost while preserving accurate target selection and action generation. This motivates examining how quantization affects GUI responses, instruction–target associations, and spatial relations among neighboring GUI elements within the interface.

Existing PTQ methods use reconstruction and activation-aware scaling \cite{gptq,awq}. Recent methods incorporate end-loss guidance, modality or token sensitivity, and action fidelity \cite{guidedquant,mbq,qig,qvla,actquant}. These advances provide useful task-aware signals, but identifying sensitive modules does not establish where finer grouping is most beneficial. For GUI agents, a sequence-averaged objective alone does not explicitly balance numerical content against the rest of the response, while local reconstruction error does not directly measure whether instruction–target and spatial relations are preserved. The grounding and action gaps in Figure~\ref{fig:intro} motivate response preservation and relation preservation for GUI quantization.

GUI quantization should preserve responses under a limited scale-byte budget and retain the instruction--vision affinities and local spatial relations needed for GUI target selection. These goals guide our choices of granularity and symmetry. \textit{Granularity} determines which weights share a scale. Groups of 64 weights (G64) adapt ranges more locally than groups of 128 (G128) but double scale metadata. Finer groups should therefore be assigned where they best preserve GUI responses. \textit{Symmetry} determines the grid within each group: asymmetric mappings adapt to offset ranges at the cost of zero-point metadata. Choosing between these grids requires assessing how they affect instruction--vision affinities and local spatial relations in GUIs. Supplementary Table~S85 (Sec.~S13) compares the accuracy of both quantization choices on the evaluated GUI tasks.

We propose \textbf{GUI-Q}, a framework with W4 weights and 16-bit activations (W4A16). Task-Sensitive Granularity Allocation (\textbf{TSGA}) estimates module sensitivity from full responses, digit-bearing tokens, and remaining tokens. It combines this sensitivity with the measured reduction in activation-weighted distortion from G128 to G64 to allocate finer groups on a per-module basis under a scale-byte budget. These label-free calibration signals use the frozen model's own responses. At the resulting group sizes, Relation-Preserving Symmetry Selection (\textbf{RPSS}) selects symmetric or asymmetric grids by their preservation of instruction--vision affinities and local spatial relations in GUIs. 

Across UI-TARS-1.5-7B, OS-Atlas-Base-7B, and InternVL3.5-8B, GUI-Q improves mean grounding TargetHit (TH) over the strongest evaluated W4 configurations by 12.55, 9.62, and 8.04 points, respectively, while reducing static peak GPU memory by 49.9--56.2\% relative to BF16. A separate AgentCPM-GUI evaluation on AndroidWorld~\cite{androidworld} improves strict success by 7.2 points. Packed four-bit model calls run 1.7--2.8$\times$ faster in our fixed-output replay. 

Our work makes the following three main contributions:
\begin{itemize}
\item A W4A16 PTQ framework for GUI agents that combines granularity allocation and symmetry selection under a scale-byte budget and a cap on asymmetric modules.
\item TSGA, which allocates finer groups based on response sensitivity and distortion reduction to preserve GUI responses under a shared budget for storing quantization scales.
\item RPSS, which selects symmetric or asymmetric grids to preserve instruction--vision affinities and local spatial relations needed to select the correct GUI target.
\end{itemize}

\begin{figure*}[!t]
\centering
\includegraphics[width=\textwidth]{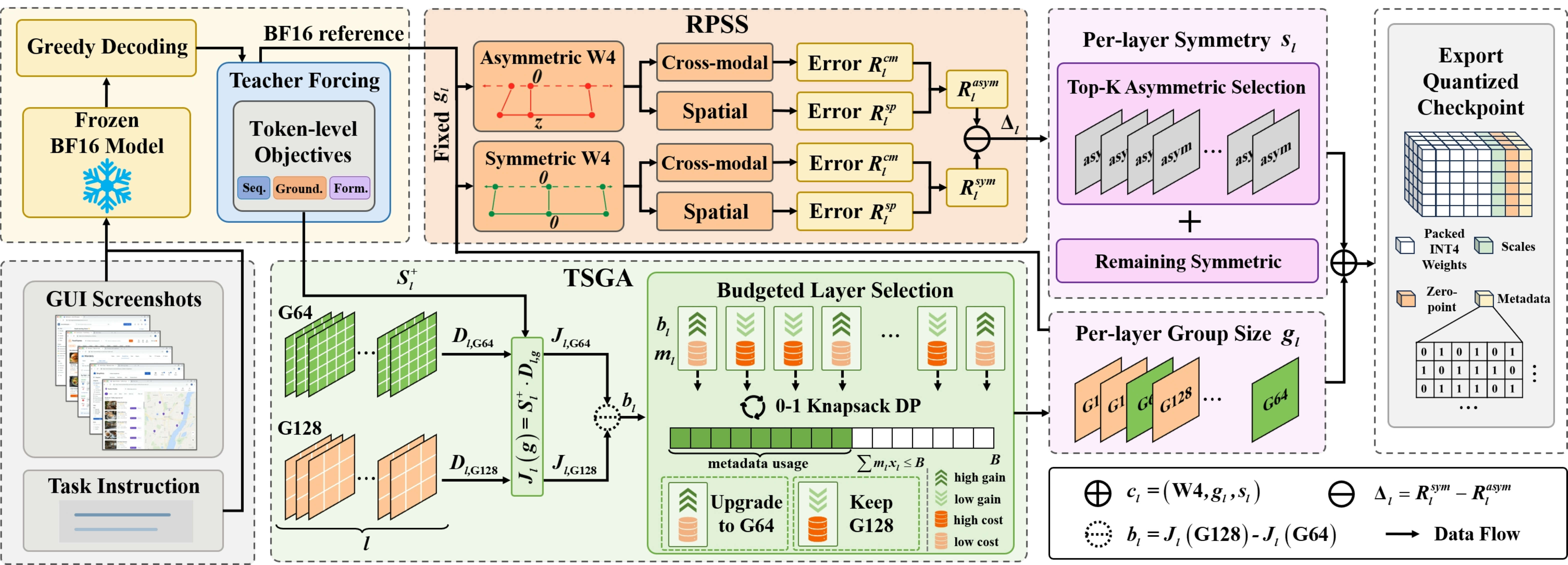}
\caption{Overview of GUI-Q: TSGA allocates task-sensitive granularity, while RPSS selects relation-preserving quantizers.}
\label{fig:pipeline}
\end{figure*}

\section{Related Work}

\paragraph{GUI understanding and agents.}
GUI systems have progressed from visual templates and interface encoders to screenshot grounding and general interaction \cite{sikuli,uibert,screenai,seeclick}. Web, mobile, and desktop benchmarks now evaluate both isolated grounding and multi-step execution \cite{screenspotpro,mind2web,webarena,visualwebarena,androidcontrol,osworld,osworldg}. UI-TARS and OS-Atlas provide representative GUI-oriented checkpoints for grounding and action generation, while InternVL3.5 provides a general multimodal checkpoint with GUI interaction capabilities. Static grounding assesses target localization, whereas closed-loop evaluation captures how action errors change later observations and task outcomes. These requirements motivate preserving target selection and action generation when quantizing GUI agents.

\paragraph{Post-training quantization.}
PTQ methods optimize weight rounding \cite{adaround,gptq}, rescale weights using activations \cite{awq}, or learn calibration transformations \cite{omniquant}. Related work also studies token-group constraints for activation quantization \cite{codtdq}. Sensitivity-based methods use gradients, curvature, perturbations, and end-loss signals \cite{hawqv2,oac,luq,guidedquant}. AutoRound is represented by the SignRound formulation and implementation family \cite{signround}. SignRoundV2 combines gradient-informed allocation with scale initialization and tuning that also apply at fixed bit width \cite{signroundv2}. Sensitivity alone does not indicate how much a module benefits from finer grouping, while distortion reduction alone does not measure its effect on GUI responses. GUI-Q combines response sensitivity with the measured G128-to-G64 distortion reduction to allocate finer groups under a shared scale-byte budget.

\paragraph{Multimodal and action-aware quantization.}
Multimodal PTQ studies dependencies across layers, modality balance, compensation for asymmetry, bit-width allocation, and token sensitivity \cite{mbq,mquant,maba,qig}. In embodied control, QVLA and ActQuant use sensitivity to action spaces or predicted actions \cite{qvla,actquant}, while QuantVLA studies scale-calibrated PTQ for vision-language-action models \cite{quantvla}. Task sensitivity and full-precision self-generated action references also appear in prior PTQ. For GUI target selection, local reconstruction error does not directly assess changes in instruction--vision affinities or local spatial relations. RPSS evaluates the preservation of these relations to select symmetric or asymmetric grids at the group sizes allocated by TSGA.

\section{Method}
\label{sec:method}

\subsection{Problem Formulation}
\label{sec:problem}
We quantize all targeted linear weights to W4, while activations and untargeted components remain BF16. We choose granularity and symmetry under their metadata constraints to preserve GUI responses and relations for target selection.

Let $f_{\theta}$ be a frozen GUI vision-language model and $\mathcal D_c=\{(I_n,u_n)\}_{n=1}^{N}$ an unlabeled calibration set of screenshot--instruction pairs. For sample $n$ and linear module $l\in\mathcal L$, $H_{l,n}$ has teacher-forced inputs at all $T_n$ response positions:
\begin{equation}
\begin{gathered}
H_{l,n}\in\mathbb R^{T_n\times d_l^{\mathrm{in}}},\qquad
W_l,\widehat W_l\in\mathbb R^{d_l^{\mathrm{out}}\times d_l^{\mathrm{in}}},\\
Y_{l,n}=H_{l,n}W_l^\top\in\mathbb R^{T_n\times d_l^{\mathrm{out}}}.
\end{gathered}
\end{equation}
Module $l$ uses $c_l=(4,g_l,a_l)$, where $g_l\in\{64,128\}$ and $a_l\in\{\mathrm{sym},\mathrm{asym}\}$ denote the group size and grid symmetry.

\paragraph{Granularity and symmetry.}
Splitting a G128 group into two G64 groups allows separate ranges, improving resolution when weight ranges differ. The two-byte scales cost 0.125 and 0.25 bits per targeted weight for G128 and G64, respectively. These supported, nested choices expose a measurable trade-off without assuming universal optimality. Sensitivity alone cannot establish the benefit of finer grouping; distortion reduction alone does not measure its effect on GUI responses. TSGA combines both signals to identify modules where the G128-to-G64 gain justifies the extra scales.

A symmetric grid ties its range to the largest absolute weight; an asymmetric grid adds a zero point for an offset range. Either can alter relative affinities between instructions and competing targets. At the assigned granularity, RPSS compares how both grids preserve instruction--vision affinities and spatial relations among nearby visual tokens.

\paragraph{Quantization mappings.}
For each input-channel group $w$ in an output row, symmetric W4 uses codes $[-7,7]$, scale $\alpha_s=\max(\max_i|w_i|/7,\epsilon_s)$, and zero point $z_s=0$. For a nonconstant asymmetric group, set $a=\min(\min_i w_i,0)$ and $b=\max(\max_i w_i,0)$. Its scale and zero point are
\begin{equation}
\begin{aligned}
\alpha_a&=\max\!\left(\frac{b-a}{15},\epsilon_s\right),\\
z_a&=\operatorname{clip}\!\left(\operatorname{rnd}\!\left(-\frac{a}{\alpha_a}\right),0,15\right).
\end{aligned}
\end{equation}
With the chosen scale $\alpha$, zero point $z$, and code bounds $q_{\min},q_{\max}$, the shared map for each weight is
\begin{equation}
\widehat w_i=\alpha\left[\operatorname{clip}\!\left(\operatorname{rnd}(w_i/\alpha)+z,q_{\min},q_{\max}\right)-z\right].
\end{equation}
Here $\operatorname{rnd}$ rounds to nearest, with ties to even. Scales are stored in BF16 before codes and zero points are computed, with $\epsilon_s=10^{-8}$. Including zero in the asymmetric range avoids collapse for one-sided groups. Supplementary Sec.~S1.3 gives constant-group, packing, and rounding rules.

\paragraph{Coupled decisions and staged selection.}
TSGA allocates group sizes under an incremental scale-byte budget $B_{\mathrm{scale}}$ relative to all-G128, using a common symmetric reference. At those group sizes, RPSS selects up to $K_{\mathrm{asym}}$ asymmetric modules $\mathcal A$ from a frozen attention-projection pool $\mathcal C$. Other modules remain symmetric. With both assignments frozen, the common rounding backend calibrates and exports the packed checkpoint (Figure~\ref{fig:pipeline}). Supplementary Sec.~S7 separately audits padding, alignment, zero points, and backend metadata. Sec.~\ref{sec:ablation} tests this staged approximation by comparing isolated RPSS scores with observed results of jointly quantized models on the downstream GUI tasks we evaluate.

\subsection{Task-Sensitive Granularity Allocation}
\label{sec:tsga}
TSGA partitions frozen-model responses to estimate task-weighted recoverable distortion. Digit-bearing tokens comprise only 17.15--20.85\% of calibration response tokens. Separate masks weight them beyond their sequence-average frequency. These shared lexical masks approximate numerical fields and remaining response structure across output formats, without parsing exact coordinates or structured actions.

For each pair, the BF16 model greedily generates
\begin{equation}
y_n^*=\operatorname{Greedy}(f_\theta,I_n,u_n).
\end{equation}
Teacher forcing under the native chat template yields
\begin{equation}
\ell_{n,k}=-\log p_\theta\left(
y_{n,k}^*\mid I_n,u_n,y_{n,<k}^*
\right).
\end{equation}
Losses exclude padding and include the end-of-sequence (EOS) token when emitted. Let $\mathcal T=\{\mathrm{seq},\mathrm{num},\mathrm{non}\}$ and $\mathcal I_n^{\mathrm{seq}}=\{1,\ldots,T_n\}$. The mask $\mathcal B_n^{\mathrm{num}}$ selects positions whose decoded token forms contain a Unicode decimal digit; $\mathcal B_n^{\mathrm{non}}$ selects the remaining positions. An empty lexical mask uses all token positions in the response:
\begin{gather}
\mathcal I_n^t=
\begin{cases}
\mathcal B_n^t, & |\mathcal B_n^t|>0,\\
\mathcal I_n^{\mathrm{seq}}, & |\mathcal B_n^t|=0,
\end{cases}
\qquad t\in\{\mathrm{num},\mathrm{non}\}.
\label{eq:mask-fallback}
\\
\mathcal L_{n,t}=
\frac{1}{|\mathcal I_n^t|}
\sum_{k\in\mathcal I_n^t}\ell_{n,k},
\qquad t\in\mathcal T.
\label{eq:masked-objective}
\end{gather}
We fix $(p_{\mathrm{seq}},p_{\mathrm{num}},p_{\mathrm{non}})=(0.30,0.50,0.20)$ before testing for all models.
With $U_{l,t,n}=\partial\mathcal L_{n,t}/\partial Y_{l,n}$, TSGA estimates module sensitivity using the following factorized proxy:
\begin{equation}
\widehat E_{l,t}=
\frac{1}{N}\sum_{n=1}^{N}
\operatorname{mean}(H_{l,n}^{\odot2})
\operatorname{mean}(U_{l,t,n}^{\odot2}).
\label{eq:gradient-energy}
\end{equation}
Response-token loss gradients propagate through visual and instruction prefixes, but $H_{l,n}$ and $U_{l,t,n}$ retain only response rows. Thus $\widehat E_{l,t}$ is a response-restricted, factorized ranking proxy. Percentile normalization makes the three objective scores comparable before weighting. Distortion measures the available benefit of changing the grid. Frozen self-generated responses provide consistent calibration targets; oracle and full-position controls test these approximations.
Ascending energy ranks with averaged ties give percentiles:
\begin{align}
\widetilde E_{l,t}
&=
\frac{\operatorname{rank}_{j\in\mathcal L}(\widehat E_{j,t})-1}
{\max(|\mathcal L|-1,1)},
\label{eq:global-percentile}\\
R_l
&=\sum_{t\in\mathcal T}p_t\widetilde E_{l,t},
\qquad
S_l^+=R_l+10^{-4}.
\label{eq:positive-sensitivity}
\end{align}
Before final rounding calibration, TSGA scores direct symmetric min--max quantization. For group size $g$, $\overline W_{l,g}^{\mathrm{sym}}$ denotes quantized weights. The activation-weighted distortion, sensitivity-weighted objective, and scale storage costs are
\begin{align}
D_{l,g}
&=
\frac{\sum_n\|H_{l,n}(W_l-\overline W_{l,g}^{\mathrm{sym}})^\top\|_F^2}
{\max\left(\sum_n\|H_{l,n}W_l^\top\|_F^2,\epsilon_D\right)},
\notag\\
J_l(g)
&=S_l^+D_{l,g},\\
C_l(g)
&=m_{\mathrm{scale}}d_l^{\mathrm{out}}
\left\lceil\frac{d_l^{\mathrm{in}}}{g}\right\rceil.
\end{align}
For the reported GUI-Q layouts, each stored scale uses $m_{\mathrm{scale}}=2$ bytes. For G64 ($x_l=1$), the predicted benefit over G128 is $b_l=\left[J_l(128)-J_l(64)\right]_+$. A high benefit requires measured distortion reduction and response sensitivity. TSGA maximizes total benefit without exceeding the byte budget for the storage of additional scales:
\begin{align}
\max_{x_l\in\{0,1\}}\quad
&\sum_{l\in\mathcal L}b_lx_l,\\
\mathrm{s.t.}\quad
&\sum_{l\in\mathcal L}
\left[C_l(64)-C_l(128)\right]x_l
\leq B_{\mathrm{scale}}.
\label{eq:tsga-knapsack}
\end{align}
The positive part assigns zero benefit when G64 does not reduce the proxy. We divide integer costs by their greatest common divisor, then solve the reduced knapsack by dynamic programming (DP), as detailed in Algorithm~\ref{alg:tsga-main}.

\begin{algorithm}[t]
\caption{Task-Sensitive Granularity Allocation}
\label{alg:tsga-main}
\begin{algorithmic}[1]
\REQUIRE $f_\theta$, $\mathcal D_c$, $\mathcal L$, $\{64,128\}$, and $B_{\mathrm{scale}}$
\ENSURE Group assignment $\{g_l\}_{l\in\mathcal L}$
\FOR{$n=1,\ldots,N$}
    \STATE Generate $y_n^*$ and build all three response masks
    \STATE Use the sequence mask for any empty lexical mask
    \STATE Teacher-force $y_n^*$ and cache $H_{l,n}$ and $U_{l,t,n}$
\ENDFOR
\STATE Compute $\widehat E_{l,t}$ and its global percentile rank
\FOR{each $l\in\mathcal L$}
    \STATE Compute $S_l^+$, $D_{l,64}$, $D_{l,128}$, $b_l$, and both scale costs
\ENDFOR
\STATE Solve Eq.~\ref{eq:tsga-knapsack} by cost-normalized DP
\STATE Set $g_l=64$ when $x_l=1$ and $g_l=128$ otherwise
\RETURN $\{g_l\}_{l\in\mathcal L}$
\end{algorithmic}
\end{algorithm}

\begin{table*}[!t]
\centering
\caption{UI-TARS-1.5-7B accuracy (\%). Bold/underline: best/second-best W4 result. BF16: full precision.}
\label{tab:main-uitars}
\setlength{\tabcolsep}{3.0pt}
\begin{tabularx}{\textwidth}{@{}l*{7}{Y}@{}}
\toprule
Method & SS-Pro TH & SS-v2 TH & AC TH & AC GA & AC SS & OSW-G TH & Mean TH \\
\midrule
BF16 & 43.01 & 87.42 & 37.70 & 75.59 & 59.77 & 29.96 & 53.47 \\
\midrule
AWQ & 22.77 & 66.19 & \underline{37.50} & 65.82 & \underline{59.18} & 4.43 & 31.13 \\
GPTQ & 18.98 & 69.89 & 25.00 & 58.20 & 46.09 & 1.42 & 30.09 \\
MABA & 0.51 & 33.96 & 15.43 & 49.02 & 44.34 & 2.84 & 12.44 \\
VLM-PTQ & 16.19 & 52.20 & 9.77 & 15.82 & 9.77 & \underline{6.03} & 24.81 \\
AutoRound & 18.41 & 43.87 & 6.64 & 6.84 & 6.25 & 4.61 & 22.29 \\
COD-TDQ & 9.42 & 64.94 & 33.98 & 68.16 & 58.40 & 3.01 & 25.79 \\
MBQ & 17.14 & 51.18 & 8.79 & 14.84 & 8.79 & 4.96 & 24.43 \\
QIG & 15.64 & 50.24 & 7.81 & 13.87 & 7.81 & 4.08 & 23.32 \\
Q-VLM & 14.17 & 50.24 & 8.79 & 12.89 & 6.84 & 3.01 & 22.47 \\
GuidedQuant & 23.53 & \underline{71.93} & 36.33 & 66.80 & 46.88 & 1.77 & 32.41 \\
QVLA & 15.62 & 65.72 & 35.94 & 71.09 & 51.17 & 4.43 & 28.59 \\
ActQuant & \underline{25.11} & 71.86 & 36.52 & \underline{72.85} & 55.66 & 2.66 & \underline{33.21} \\
\textbf{GUI-Q} & \textbf{35.62} & \textbf{85.91} & \textbf{37.70} & \textbf{74.41} & \textbf{59.57} & \textbf{15.74} & \textbf{45.76} \\
\bottomrule
\end{tabularx}
\end{table*}

\subsection{Relation-Preserving Symmetry Selection}
\label{sec:rpss}
RPSS selects grids using relations needed for target selection. Instruction--vision affinities link screenshot features to the request; spatially weighted visual similarities describe local feature relations. Preserving both complements response-sensitive allocation. After TSGA fixes $\{g_l\}$, RPSS evaluates a frozen 16-module attention pool spanning early q projections and k/v/o projections at three depths. Attention projections directly affect these relations, while the frozen pool bounds profiling cost. Supplementary Sec.~S2.8 details pools, tie orders, and how visual tokens map to screenshots.

For candidate $a=(b,p)\in\mathcal C$, RPSS quantizes only $a$ at $g_a$ and retains BF16 elsewhere. This estimates the grid's marginal gain in relation preservation at its assigned granularity. It omits interactions among jointly quantized projections. Preserving relative affinities cannot guarantee unchanged downstream actions, because they do not constrain every feature direction used by the final readout. We measure features at block $b$ after attention and residual addition but before the multilayer perceptron (MLP), then apply $\ell_2$ normalization. Text features use only user-instruction positions. Visual features use visible screenshot tokens after the frozen projector and native token compression. Visual-token positions are normalized centroids of their source patches.
Let $T_{a,n}^r\in\mathbb R^{M_n\times d}$ and $V_{a,n}^r\in\mathbb R^{V_n\times d}$ be text and visual features for $r\in\{\mathrm{fp},\mathrm{sym},\mathrm{asym}\}$. Their normalized dot products define the affinity of each token pair:
\begin{equation}
A_{a,n,ij}^r=
\frac{\langle T_{a,n,i}^r,V_{a,n,j}^r\rangle}{\tau}.
\label{eq:affinity}
\end{equation}
We compute $P_{a,n}^r=\operatorname{LogSinkhorn}(A_{a,n}^r,\mu_n,\nu_n)$, where $\mu_n,\nu_n\geq0$, $\mathbf 1^\top\mu_n=\mathbf 1^\top\nu_n=1$, $P_{a,n}^r\mathbf 1=\mu_n$, and $(P_{a,n}^r)^\top\mathbf 1=\nu_n$. Log-domain normalization avoids unstable exponentiation and yields positive transport entries. We define the cross-modal discrepancy using these couplings:
\begin{equation}
\mathcal L_{a,n}^{\mathrm{cm}}(r)=
\sum_{i=1}^{M_n}\sum_{j=1}^{V_n}
P_{a,n,ij}^{\mathrm{fp}}
\log\frac{P_{a,n,ij}^{\mathrm{fp}}}{P_{a,n,ij}^{r}}.
\end{equation}

Let $\rho_{n, i}\in[0,1]^2$ be the visual-token positions defined above.
We define the spatial relation and its discrepancy, then use
their combined score to select asymmetric modules:
\begin{align}
G_{a,n,ij}^r
&=
\langle V_{a,n,i}^r,V_{a,n,j}^r\rangle
\notag\\*
&\quad{}\times\exp\left(
-\frac{\|\rho_{n,i}-\rho_{n,j}\|_2^2}{2\sigma^2}
\right),\\
\mathcal L_{a,n}^{\mathrm{sp}}(r)
&=
\frac{\|G_{a,n}^{\mathrm{fp}}-G_{a,n}^{r}\|_F^2}
{\|G_{a,n}^{\mathrm{fp}}\|_F^2+\epsilon_G},\\
\mathcal R_{a,n}(r)
&=
\mathcal L_{a,n}^{\mathrm{cm}}(r)
+\gamma\mathcal L_{a,n}^{\mathrm{sp}}(r),\\
\Delta_a
&=
\frac{1}{N}\sum_{n=1}^{N}
\left[
\mathcal R_{a,n}(\mathrm{sym})
-\mathcal R_{a,n}(\mathrm{asym})
\right],\\*
\mathcal A
&=
\arg\max_{\substack{\mathcal S\subseteq\mathcal C\\
|\mathcal S|\leq K_{\mathrm{asym}}}}
\sum_{a\in\mathcal S}[\Delta_a]_+.
\end{align}
RPSS selects up to $K_{\mathrm{asym}}$ candidates with positive scores; if fewer are positive, it selects fewer modules. Ties follow the frozen candidate order. We use $\tau=0.1$, $\sigma=0.25$, $\gamma=1$, and $\epsilon_D=\epsilon_G=10^{-8}$. These marginal scores require validation against the observed outcomes of the final jointly quantized model on the GUI tasks we evaluate.

\paragraph{Final rounding calibration.}
After allocation, the common backend optimizes only zero-initialized rounding offsets in $[-1,1]$. Weights, BF16-stored scales, zero points, and constant groups remain fixed. It uses all valid teacher-forced positions to minimize normalized activation-weighted reconstruction error over 50 Adam steps. It uses 32-bit floating-point (FP32) arithmetic and straight-through rounding. We export hard codes from the earliest of the 51 iterates attaining the lowest error over the full calibration cache.

\begin{figure*}[!t]
\centering
\includegraphics[width=\textwidth]{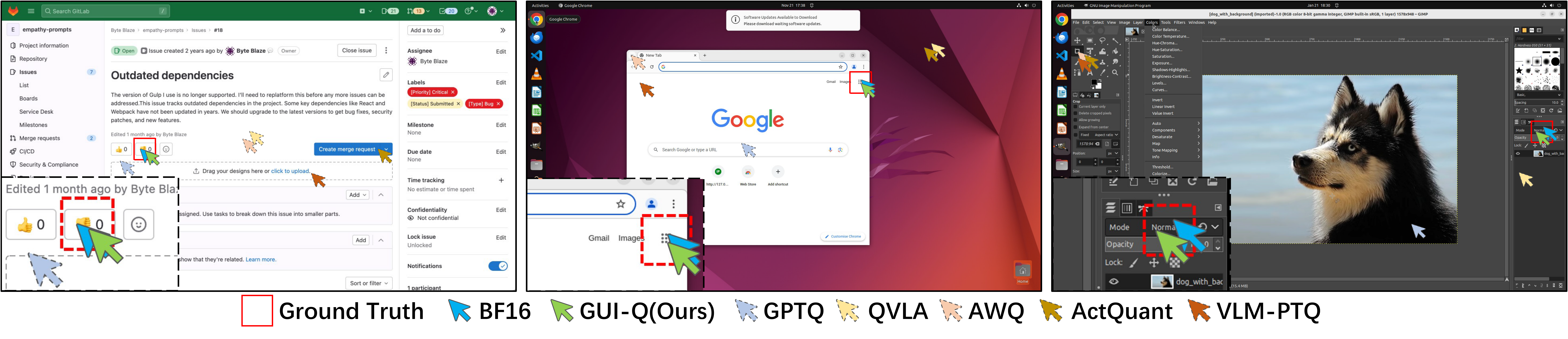}
\caption{Qualitative comparison on GUI tasks. GUI-Q correctly localizes the requested control in all three cases.}
\label{fig:qualitative-analysis}
\end{figure*}

\section{Experiments}
\label{sec:experiments}

\subsection{Experimental Setup}\label{sec:setup}
\paragraph{Models and evaluation metrics.}
We evaluate UI-TARS-1.5-7B, OS-Atlas-Base-7B, and InternVL3.5-8B on ScreenSpot-Pro (SS-Pro), ScreenSpot-v2 (SS-v2), AndroidControl (AC), and OSWorld-G (OSW-G), where their output formats are supported. We use official TH, Grounding Accuracy (GA), and StepSuccess (SS). Mean TH averages the three shared grounding benchmarks; AndroidControl endpoints are reported separately. Closed-loop AndroidWorld uses 250 frozen task--reset pairs. Strict success requires both a reward above $0.5$ and an explicit termination action.

\paragraph{PTQ protocol and baselines.}
All methods quantize decoder q/k/v/o and gate/up/down projections to W4, retain BF16 elsewhere, and share the evaluation interface. We compare AWQ, GPTQ, AutoRound, VLM-PTQ, MABA, COD-TDQ, MBQ, QIG, Q-VLM, and allocation/action-aware transfers. Transfers share calibration data, 384 backward passes, the rounding backend, and the asymmetry cap. GUI-Q uses 128 unlabeled calibration examples, a 256-token response cap, and 50 rounding iterations. Calibration, development, static tests, and AndroidWorld are disjoint. Lexical weights are selected on a separate annotated 512-example OS-Atlas development split and frozen before testing. Only calibration is label-free. Static W4 results average five paired seeds on fixed test examples; BF16 is deterministic. Each comparison fixes prompts, image processing, parsers, stopping rules, and official evaluators. GuidedQuant-, QVLA-, and ActQuant-inspired transfers implement allocation or action signals in this W4 scope, rather than reproducing every upstream recipe. A separate retained-objective cohort helps separate criterion gains from adaptation effects. Supplementary Secs.~S2.2--S2.9 specify seeds, baseline settings, hyperparameters, manifests, data splits, and the hardware.

Configuration and retained-objective controls use a separate cohort: 128 cases for each of 12 static model--workload pairs and 250 AndroidWorld cases, with five paired seed slots per case. Here $G_3$ averages TH on SS-Pro, SS-v2, and OSW-G (Supplementary Secs.~S12--S13). In the primary static models, TSGA assigns G64 to 56/196, 20/196, and 81/252 targets; RPSS selects 5, 5, and 6 asymmetric projections, respectively. Remaining target modules retain symmetric grids or G128 as assigned. Supplementary Secs.~S2.6 and S6.2 give each model's byte ceilings and group assignments.

\subsection{Baseline Comparison}
\label{sec:main-comparison}

Table~\ref{tab:main-uitars} shows uneven grounding and action retention among the baselines. AWQ and GPTQ reach 31.13 and 30.09 Mean TH but only 4.43 and 1.42 OSW-G TH. VLM-PTQ leads the tested PTQ baselines on OSW-G yet reaches only 9.77 AndroidControl SS. GUI-Q attains 45.76 Mean TH and 37.70/74.41/59.57 AndroidControl TH/GA/SS. The controls below test whether these gains depend on the criteria. These metrics differ: a valid action schema does not guarantee the correct target, and an accurate isolated click does not establish trajectory success. We report AndroidControl, closed-loop success, and the mean grounding accuracy.

On the separate cohort, GUI-Q exceeds retained-objective GuidedQuant and Action-Mixed-Fisher controls by 4.53 and 3.70 $G_3$ points on UI-TARS, and 5.36 and 4.08 strict-success points on AgentCPM-GUI. The respective 95\% case-cluster intervals are $[3.49,5.57]$ and $[2.71,4.69]$ points for UI-TARS, and $[3.84,6.88]$ and $[2.64,5.52]$ for AndroidWorld. These contrasts support gains beyond common-backend transfer controls. Supplementary Sec.~S12 provides all endpoints and case-cluster intervals for this cohort.

\paragraph{Closed-loop AndroidWorld.}
On the same 250 frozen task--reset pairs, AgentCPM-GUI with GUI-Q reaches 31.2\% strict success, versus 24.0\% for the strongest evaluated QVLA-inspired transfer and 38.4\% for BF16 (Table~\ref{tab:main-agent}). All actions are parseable; ``Reward'' counts reward successes. The 7.2-point gain reflects 34 GUI-Q-only versus 16 QVLA-only successes (Holm-adjusted exact McNemar $p=0.01535$). The unadjusted paired-bootstrap 95\% interval is $[1.6,12.8]$ points; the Bonferroni--Wald simultaneous 95\% interval is $[-0.32,14.72]$. Supplementary Sec.~S4.2 explains how the paired tests and interval methods differ.

On the same tasks, five seeds yield gains of 7.2--7.6 points (mean: 7.28). A disjoint 250-pair replication yields 30.8\% versus 23.2\% ($+7.6$ points), with 33 GUI-Q-only versus 14 QVLA-only successes (unadjusted exact McNemar $p=0.00794$). Repeated seeds assess configuration stability, without creating independent task samples. Exploratory errors suggest fewer coordinate, neighboring-target, and action-field failures, with differing failure denominators. Supplementary Sec.~S8 reports independent replication outcomes, paired seed results, task strata, and error counts.

\begin{table}[!t]
\centering
\caption{AgentCPM-GUI on 250 AndroidWorld episodes.}
\label{tab:main-agent}
\setlength{\tabcolsep}{3.4pt}
{\small
\begin{tabular}{cccccc}
\toprule
Method & Strict & Reward & Method & Strict & Reward \\
\midrule
BF16        & 38.4\% & 96 &
AWQ         & 20.4\% & 54 \\
GPTQ        & 17.6\% & 47 &
GuidedQuant & 21.6\% & 57 \\
ActQuant & 23.2\% & 61 &
VLM-PTQ     & 15.6\% & 42 \\
QVLA & \underline{24.0\%} & \underline{63} &
\textbf{GUI-Q} & \textbf{31.2\%} & \textbf{81} \\
\bottomrule
\end{tabular}
}
\end{table}

\begin{table}[!t]
\centering
\caption{Component-removal ablation on UI-TARS.}
\label{tab:component-ablation}
\setlength{\tabcolsep}{2.5pt}
{\small
\begin{tabular*}{\columnwidth}
{@{\extracolsep{\fill}}lccccc@{}}
\toprule
Configuration
& SS-Pro
& SS-v2
& \mbox{OSW-G}
& AC GA
& AC SS \\
\midrule
\textbf{GUI-Q}
& \textbf{35.62}
& \textbf{85.91}
& \textbf{15.74}
& \textbf{74.41}
& \textbf{59.57} \\
\textit{w/o TSGA}
& 11.57
& 65.57
& 13.83
& 55.86
& 40.43 \\
\textit{w/o RPSS}
& 14.04
& 71.23
& 15.43
& 73.24
& 57.23 \\
\bottomrule
\end{tabular*}
}
\end{table}

\begin{table*}[!t]
\centering
\caption{Criterion-replacement ablation on UI-TARS.}
\label{tab:criterion-ablation}
\setlength{\tabcolsep}{3.0pt}
\begin{tabular*}{\textwidth}
{@{\extracolsep{\fill}}llcccccc@{}}
\toprule
Allocation objective
& Symmetry criterion
& SS-Pro
& SS-v2
& OSW-G
& Mean TH
& AC GA
& AC SS \\
\midrule
\textbf{GUI lexical proxy}
& \textbf{GUI relation}
& \textbf{35.62}
& \textbf{85.91}
& \textbf{15.74}
& \textbf{45.76}
& \textbf{74.41}
& \textbf{59.57} \\
& Attention MSE
& 33.19
& 84.10
& 12.94
& 43.41
& 73.24
& 58.59 \\
Generic end loss
& GUI relation
& 34.00
& 84.75
& 13.65
& 44.13
& 73.83
& 58.98 \\
& Attention MSE
& 28.61
& 74.43
& 10.60
& 37.88
& 65.00
& 51.88 \\
\bottomrule
\end{tabular*}
\end{table*}

\subsection{Ablation Study}
\label{sec:ablation}
Tables~\ref{tab:component-ablation} and~\ref{tab:criterion-ablation} compare component removal and criterion replacement with identical calibration data, decoder targets, W4 payload, scale-byte ceilings, asymmetry caps, rounding backend, and evaluation protocol. Stage removal measures its full contribution and can change realized metadata bytes. Matched-byte controls and criterion replacements assess selection quality under the same storage constraints.

\paragraph{Why these configuration axes?}
Supplementary Table~S85 (Sec.~S13) tests granularity and symmetry on the separate cohort. G64-sym improves $G_3$ over G128-sym by 7.09, 7.08, and 7.97 points on UI-TARS, OS-Atlas, and InternVL, respectively. RPSS adds 5.99, 6.98, and 6.46 $G_3$ points at fixed G128 and also improves all three $G_3$ scores at fixed G64. Finer grouping thus does not remove symmetry selection's benefit. Relative to uniform G64 plus RPSS, GUI-Q changes $G_3$ by $+0.42$, $+0.26$, and $-0.16$ points while using less decoder storage. These closely spaced point estimates support selective allocation under a metadata constraint within this cohort. They establish neither uniform accuracy dominance nor statistical equivalence on the evaluated tasks.


\paragraph{Component removal.}
Uniform G128 with recomputed RPSS (\textit{w/o TSGA}) lowers Mean TH by 15.44 points and AndroidControl GA/SS by 18.55/19.14 points, removing both task-aware allocation and its extra scales. Keeping TSGA groups but making all modules symmetric (\textit{w/o RPSS}) lowers Mean TH by 12.20 points, with smaller action changes. The largest RPSS-related reductions occur on ScreenSpot-Pro and ScreenSpot-v2, consistent with its cross-modal and local-spatial criteria. Supplementary Sec.~S5.1 reports endpoint contrasts and storage for each removed stage.

Matched-byte controls isolate selection quality. Let $G_4$ average the three shared grounding THs and AndroidControl TH. Selecting $q_0$--$q_4$ gives 40.54 versus 43.74 for RPSS's $q_0,q_1,q_2,q_3,q_6$: a paired gain of 3.21 points (nominal five-seed 95\% interval: $[2.91,3.51]$). Cross-modal-only and spatial-only criteria reach 40.98 and 40.07; matched-byte random selection reaches 33.79. Weight mean squared error (MSE) and raw affinity MSE reach 35.58 and 40.39. These comparisons support the combined relation criterion. Supplementary Sec.~S5.6 gives the exact selections and results for all five seeds used for random selection.

\paragraph{Criterion specificity and interaction.}
Against generic end loss with attention MSE, the GUI allocation criterion alone adds 5.53 Mean TH points, the relation criterion alone 6.25, and both 7.88 (Table~\ref{tab:criterion-ablation}). The conditional lexical/relation gains are 1.63/2.35 points, with Bonferroni-adjusted five-seed 95\% intervals $[0.40,2.86]$/$[1.28,3.41]$. Their $-3.90$-point interaction ($[-5.64,-2.17]$) indicates overlapping benefits; both conditional gains remain positive. These intervals quantify seed variation on fixed test sets. Supplementary Sec.~S5.2 fully analyzes the main effects and interaction.

\paragraph{Ranking diagnostics.}
Before RPSS, matched-byte all-symmetric controls yield $G_4=33.67$ for TSGA, versus 29.82 for random allocation and 30.71 for distortion alone. TSGA attains 0.57 oracle-ranking correlation and captures 84\% of the oracle benefit in 29.8 minutes. Exact sequence Fisher, Hutchinson curvature, and oracle search reach 34.02, 34.10, and 34.46 in 42.5, 83.4, and 173.6 minutes, respectively. These controls show the factorized score's profiling--accuracy trade-off. Including prefix positions changes Mean TH from 45.76 to 45.86 at unchanged storage. RPSS attains 0.72 oracle-ranking correlation, 82\% overlap, and 84\% positive precision. Supplementary Secs.~S5.3 and S5.6 report ranking and compute diagnostics across random seeds and repeated samples drawn from the calibration data.

\paragraph{Joint-W4 validation.}
Measured UI-TARS pair corrections for the deployed five-projection set total $-1.39$ $G_4$ points (12.07\% of summed singleton gains). The second-order prediction is 43.83 versus 43.74 measured, but overpredicts fixed early-q performance by 3.22 points. Direct joint evaluation remains necessary. A 21-set equal-byte search finds a single swap 0.20 points above RPSS, with a nominal five-seed 95\% interval of $[-0.20,0.61]$; this small improvement is not statistically resolved. Restricted OS-Atlas and InternVL searches improve Mean TH by at most 0.14 and 0.12 points. Supplementary Sec.~S10 gives all 16 singletons, 120 measured pairs, and cross-model searches. Among 112, 112, and 144 eligible projections, the frozen pool covers 4/5, 4/5, and 5/6 of the full-search top-$K$ on UI-TARS, OS-Atlas, and InternVL. Full search improves the reported aggregates by only 0.13/0.13/0.12 points in these checks, supporting the practical pool while leaving its coverage task-dependent. Exact joint enumeration tests 1,368/2,394 configurations at UI-TARS scale-byte ceilings of 12.80/21.76 MB. It exceeds sequential selection by 0.18/0.14 $G_4$ points. Supplementary Sec.~S6.4 details the search space at each budget.

\paragraph{Lexical-mask and calibration robustness.}
On UI-TARS, the weighted lexical proxy gives 45.76 Mean TH, versus 45.08 with uniform weights, 45.59 with a coordinate parser, and 45.93 with an action-schema parser at the same budget ceiling. With matched compute, sequence-, numeric-, and nonnumeric-only objectives give 44.08, 44.69, and 43.72; matched random masks give 40.04. Audited digit-mask precision/recall is 73.29/88.52\%, with 5.47\% fallback. Weight perturbations of $\pm0.05$ give 45.45--45.63 Mean TH. Fixed versus uniform weights give 35.36/34.69 on OS-Atlas and 49.45/48.95 on InternVL, supporting transfer of the weights. On AgentCPM-GUI, this comparison yields 78/250 versus 74/250 strict successes, with nine weighted-only and five uniform-only successes (exact McNemar $p=0.42395$). The weights are stable across tested settings, but their closed-loop advantage over uniform weights remains statistically unresolved. Supplementary Secs.~S5.4--S5.5 and S9 report token audits and tests of the lexical weights.

Using 64, 128, and 256 calibration examples gives 44.92, 45.76, and 45.86 Mean TH, respectively. Profiling takes 27.97, 54.43, and 107.61 minutes. At $N=128$, single-domain calibration gives 44.96--45.41 versus 45.76 for mixed domains. These results support 128 mixed-domain examples as an accuracy--cost compromise. Supplementary Sec.~S6.1 reports all calibration sizes and source domains.

\paragraph{RPSS operating settings.}
Sweeps of temperature, relation weight, and radius, each varied alone, yield $G_4=43.09$--$43.74$. With 10, 20, and 30 Sinkhorn iterations, $G_4$ is 43.54, 43.74, and 43.73, respectively. Caps of 3, 5, and 7 asymmetric modules give 42.91, 43.74, and 43.63. More than 20 iterations or five asymmetric projections does not improve the reported UI-TARS score. Supplementary Sec.~S6.3 lists the full parameter sweeps and profiling costs.

\subsection{Qualitative and Visualization Analysis}
\label{sec:qualitative}

Figure~\ref{fig:qualitative-analysis} shows adjacent-target, small-object, and semantic-distractor errors. GPTQ approaches GitLab's reaction controls but misses the thumbs-down button. The small Google Apps grid icon competes with the Chrome menu and desktop clutter. In GIMP, expanded menus distract from the narrow layer-mode dropdown. GUI-Q selects all three intended controls. These examples illustrate the motivation; controlled ablations provide quantitative evidence. Supplementary Figure~S1 and Sec.~S3.3 give further examples of GUI targets.

\subsection{Cross-Model Generalization}
\label{sec:cross-model}
Across the three shared grounding benchmarks, UI-TARS achieves 53.47 Mean TH in BF16, 33.21 with the ActQuant-inspired transfer, and 45.76 with GUI-Q. OS-Atlas scores 39.22, 25.74, and 35.36 under the same settings. InternVL scores 50.21 in BF16, 41.41 with its strongest baseline AWQ, and 49.45 with GUI-Q. The gains over W4 baselines are 12.55, 9.62, and 8.04 points, with respective gaps to BF16 of 7.71, 3.86, and 0.76 points. GUI-Q improves all three checkpoints, but recovery of full-precision accuracy varies.

GUI-Q's SS-Pro/SS-v2/OSW-G TH is 11.76/74.03/20.28 on OS-Atlas and 22.45/87.74/38.16 on InternVL. On OS-Atlas, it exceeds the best tested PTQ baseline on each benchmark. InternVL AndroidControl SS is 69.77 versus 62.89 for BF16 and 56.99 for AWQ. Supplementary Sec.~S3 provides per-dataset rates and hit totals. Dynamic-resolution checks retain these gains (Supplementary Sec.~S2.8): Mean TH is 51.72/41.02 on UI-TARS and 37.93/29.88 on OS-Atlas for GUI-Q and the listed comparator, in that order.

On the separate retained-objective cohort, GUI-Q exceeds Action-Mixed-Fisher by 3.28 $G_3$ points on OS-Atlas and 4.06 on InternVL. The respective 95\% case-cluster intervals are $[2.40,4.22]$ and $[3.13,5.05]$ (Supplementary Sec.~S12). This comparison tests whether the gains persist when the baseline objective is retained. These gains are measured within this cohort and should therefore be read separately from those on the main static tests.

\subsection{Efficiency Analysis}
\label{sec:efficiency}
We report one-time configuration cost, static peak allocation, and packed-W4 model-call latency. With 128 calibration examples, UI-TARS configuration takes 29.80 minutes for TSGA and 24.63 for RPSS, peaking at 15.43 GiB; DP takes 0.138 seconds. Static exports use 4.166 bits per targeted weight on UI-TARS, 4.145 on OS-Atlas, and 4.172 on InternVL, including quantizer metadata. Supplementary Sec.~S7 lists scale, zero-point, descriptor, BF16-rest, and serialization bytes. The AgentCPM-GUI export is audited separately in the same section as the static exports.

GUI-Q lowers peak allocation from 16.62 to 8.32 GiB on UI-TARS, 15.75 to 6.93 GiB on OS-Atlas, and 16.67 to 7.29 GiB on InternVL: reductions of 49.9\%, 56.0\%, and 56.2\%. Fixed-BF16-output replay from vision encoding through decoding gives respective speedups of 1.92--2.80$\times$, 1.70--2.80$\times$, and 1.70--2.24$\times$. AndroidWorld first-decision replay over 250 inputs falls from 792.04 to 597.08 ms (1.33$\times$), with 56.67\% less peak memory. These W4-versus-BF16 gains concern packed model calls; within-W4 configuration controls show much smaller timing differences. Fixing BF16-reference output lengths separates execution speed from action changes and early stopping. Logical dispatch coverage and reduced-dimensional numerical checks validate execution for tested operator shapes and numerical tolerances. Supplementary Secs.~S7.6 and S14 report workload timings, paired-bootstrap intervals, dispatch coverage, and numerical checks of the execution path. The AndroidWorld replay speedup has a paired-bootstrap 95\% interval of $[1.29,1.36]\times$; the 2.80$\times$ UI-TARS OSWorld-G speedup has a 95\% interval of $[2.67,2.92]\times$. These intervals support faster model calls under fixed output lengths. Environment execution is excluded. The measured speedup therefore concerns model calls and does not describe how long an agent takes to complete a GUI task in real environments.

\section{Conclusion}

\looseness=-1
GUI-Q combines TSGA and RPSS for decoder-only W4A16 quantization. TSGA allocates finer groups using response sensitivity and distortion reduction to preserve GUI responses under a scale-byte budget. RPSS selects symmetric or asymmetric grids to preserve instruction--vision affinities and local spatial relations needed for target selection. Calibration uses the frozen model's own responses without task labels. GUI-Q improves grounding over the evaluated W4 baselines and increases task success in a separate closed-loop evaluation. Static peak GPU memory falls by 49.9--56.2\% relative to BF16. Packed model calls run 1.7--2.8$\times$ faster when replaying the same fixed outputs.

\section*{Acknowledgements}
This work was supported in part by the Tianjin Science and Technology Major Project (No. 25ZXRGGX00120), in part by the National Natural Science Foundation of China (No. 62576176), and in part by the Fundamental Research Funds for the Central Universities (Nankai University, No. 070-63253235). The computational resources were supported by the Supercomputing Center of Nankai University (NKSC).

\section*{LLM Usage Statement}
ChatGPT (OpenAI) was used to assist with language polishing and LaTeX formatting adjustments in the main paper and supplementary material. The core research ideas and methodological design were developed independently by the authors without LLM involvement. The authors reviewed all AI-assisted content and take full responsibility for the manuscript.

{
    \small
    \IfFileExists{references.bib}{%
        \bibliographystyle{ieeenat_fullname}
        \bibliography{references}%
    }{%
        \input{references_recovered.bbl}%
    }
}

\end{document}